# Strategies for Resource Allocation of Two Competing Companies using Genetic Algorithm

**Wing Keung Cheung and Kwok Yip Szeto**
Department of Physics
Hong Kong University of Science and Technology
Clear Water Bay, Hong Kong SAR, China
e-mail: phszeto@ust.hk

**Abstract**: *We investigate various strategic locations of shops in shopping malls in a metropolis with the aim of finding the best strategy for final dominance of market share by a company in a competing environment. The problem is posed in the context of two competing supermarket chains in a metropolis, described in the framework of the two-dimensional Ising model. Evolutionary Algorithm is used to encode the ensemble of initial configurations and Monte Carlo method is used to evolve the pattern. Numerical simulation indicates that initial patterns with certain topological properties do evolve faster to market dominance. The description of these topological properties is given and suggestions are made on the initial pattern so as to evolve faster to market dominance.*

## Introduction

An interesting problem in econophyiscs addresses the problem of resource allocation with the objective of maximizing the market share. The resource of each company is limited so that the problem of resource allocation is strategically important for the future evolution of its market share. The recent model of Szeto and Kong [1] considers two supermarket chains in a metropolis. The resource of each company is not sufficient to open shops in all the malls and each mall can accommodate only one shop. The problem for each company is to allocate the resource for shops so as to maximize the market share in the future, taking into consideration of shops changing hands and/or relocation by the management of each company. This rather complex situation requires the a mathematical description of the competition and cooperation of companies. Previous studies [1-3] on this model addressed the econophysics problem in the context of multi-agent systems. The basic tools used in our analysis are techniques in statistical mechanics and Monte Carlo simulation. Thus the primitive approach of solving this problem is (1) generate all the possible configurations with no violation of the resources constraint, (2) perform Monte Carlo simulation of the model based on statistical mechanics and observe the evolution of the pattern of the market shares of these two companies. This approach however does not solve the problem of finding a good initial configuration of resource allocation satisfactorily, since the number of possible initial configurations is increasing exponentially with the number of shopping malls. A better way to address the selection of initial configuration of shops allocation is to reformulate our problem as an optimization problem and then solve it by various techniques of optimization. Here we will use genetic algorithm [4]. In this paper, we first introduce the model based on statistical mechanics, with well defined constraints on resources and the associated switching dynamics. We then describe the implementation of the evolutionary algorithm and Monte Carlo simulation. Finally, we make use of the topological analysis to analyze the optimal configuration.

## The Model and Resources Constraints

The ferromagnetic Ising model in statistical physics is employed to study agent-based interacting companies. This model has been extensively studied in physics of magnetism [5] and recently used in econophysics [1,6]. It considers two spin orientations (up or down) (or two companies P for Park and Shop (Blue) and W for Welcome (Red) in Hong Kong) on the lattice, and describes the interaction between two nearest neighbor spins on the lattice (with site labeled by *(i,j)* with energy (E),

$$E = \sum_{u=P}^{W}\sum_{v=P}^{W}\sum_{<i,j>} J_{ij}^{uv} x_i^u x_j^v \quad \text{with} \quad J_{ij}^{uv} = \begin{cases} 0 & \text{if} & i = j \\ -1 & \text{if} & i \neq j \text{ and } u = v \\ 1 & \text{if} & i \neq j \text{ and } u \neq v \end{cases} \quad (1)$$

Here $J_{ij}^{uv}$ is the interaction between two nearest-neighboring agents, $x_i^u$ is the binary vector represents the occupancy of company *u* on the shopping mall located at site *i*. Since we have two companies, we can use the Ising model in statistical mechanics and assign two spin orientations to represent two competing companies, and the sites in a hexagonal lattice represent shopping malls. Only one shop is available for each mall. The analogy of market competition is shown in *Figure 1*. Generalization of more than two companies and on other non-regular point patterns have been discussed elsewhere but here we focus our attention on strategy selection in resource allocation for the simplest model of two-companies on hexagonal lattice.

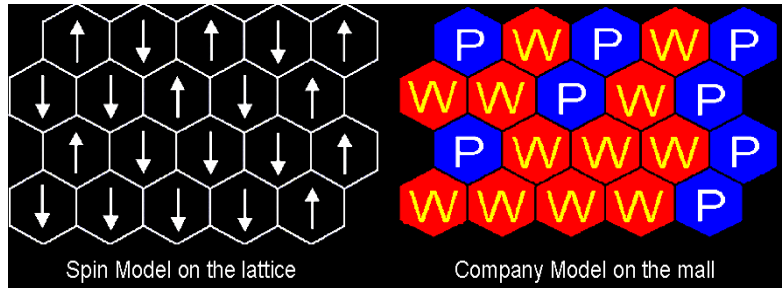

*Figure 1 Analogy from Spin model to Market Competition model*

For simplicity, we assume only one shop can be open in each shopping mall. The resources constraint of the initial configuration is the same for each company, 50% of the total number of shopping malls. The total number of shopping malls available is 100. Each company must use up all the resources available.

## Switching Dynamics

We consider the One-shop switching process. The transition probability of switching shop is given by,

$$P(i \rightarrow f) = \frac{1}{1 + e^{2\beta(\Delta E_i)}} \quad (2)$$

where $\Delta E_i$ is the change in energy when the shop switches from P to W or W to P in mall *i*, The temperature T represents the noise factor, $\beta = 1/(kT)$ the inverse temperature, and the Boltzmann constant *k=1*. Discussion of the interpretation of this model in economics can be found in Ref. [1].

## Evolutionary Algorithm and Monte Carlo Simulation

The strategy of shops allocation can be represented by a binary chromosome. In this paper, we have 100 malls in our metropolis so that the length of the chromosome is 100. We use "0" to represent shop occupancy of company P, and "1" to represent shop occupancy of company W. To generate a population of initial configurations, we first prepare a chromosome with 50% "0" and 50% "1", and then randomly swap the position of loci. We repeat this procedure until sufficient chromosomes are generated and that form the initial population of our genetic algorithm. Monte Carlo simulation is then performed to evolve the pattern according to the switching dynamics. The number of MC steps required depends on the system reaching the equilibrium state. This requirement is discussed in the literature [7,8]. We set the MC steps to be 50 such that the system satisfies the requirement. This switching mechanism is shown in *Figure 2*.

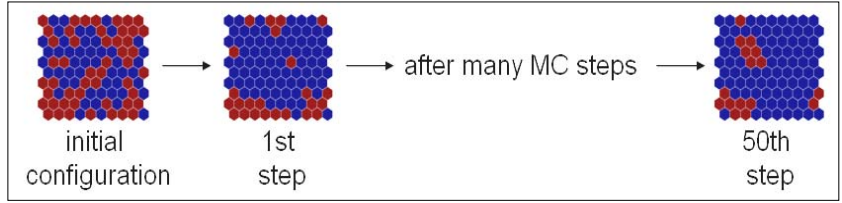

*Figure 2 Monte Carlo Simulation according to the switching dynamics*

One should note that the switching dynamics depends on temperature. In the context of econophysics, the switching process represents the change of loyalty of a particular agent from one company to another. Temperature, or the effect of thermal noise, corresponds for example to the effects of promotion to an agent in the process of competing for customers to switch their loyalty in a noisy environment. The average of market share is finally computed and served as the fitness function of the chromosome in the population pool of initial configurations. (Note that a given initial configuration corresponds to a given strategy of resource allocation of the company on the hexagonal network.) After the fitness is computed for each chromosome of the population, selection based on fitness is performed. Regeneration of the pool of chromosomes for the non-survivors is achieved by mutation of the survivors. Mutation operator is defined to switch the locus of a chromosome. The difficulty of implementing the mutation process is the violation of resources constraint. This problem can be resolved by using repair: when the chromosome violates the resources constraint, we mutate the chromosome till the resources constraint is satisfied. By repeating the whole process, we can determine the optimal initial configuration with the highest average market share. *Figure 3* shows the implementation of our algorithm.

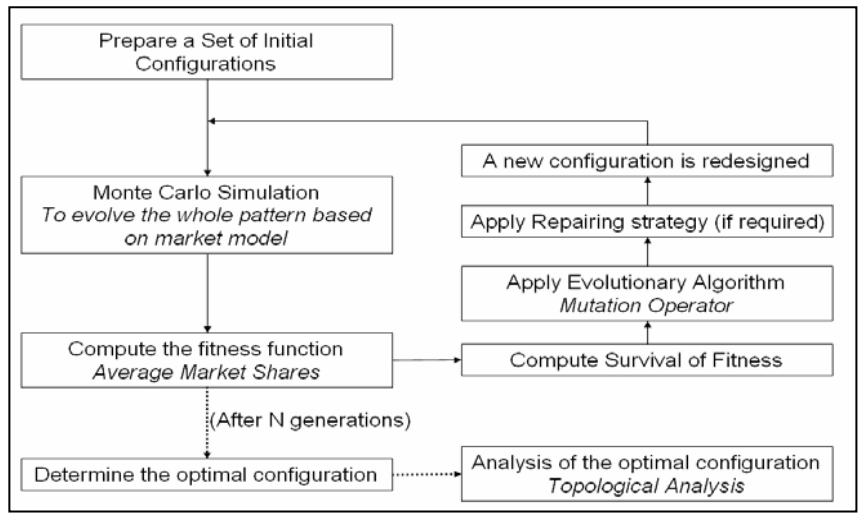

*Figure 3 Implementation of the whole algorithm*

The optimal configuration is analyzed for its topological properties, which is a common technique in high dimensional network analysis [9]. Here we use the connectivity distribution, which is the distribution of number of edges connected to the neighboring shops from a mall, as the quantitative measurement of topological properties.

# Result and Discussion

Genetic algorithm is a good optimization algorithm with great flexibility and efficiency. The setting of our genetic algorithm is shown in *Table 1*. The corresponding setting in Monte Carlo simulation is also shown in *Table 2*.

*Table 1: The parameters in genetic algorithm*

| Parameter | Value |
|---|---|
| Population Size | 50 |
| Probability of mutation | 0.03 |
| Number of generation | 100 |

*Table 2: The parameters in Monte Carlo Simulation*

| Parameter | Value |
|---|---|
| Temperature | 1, 3, 4 |
| MC Step | 50 |

We focus on the evolution of the pattern at low noise level (temperature). It is due to the fact that market competition consists of two distinct phases. One is the cluster (ferromagnetic) phase and the other is the random (paramagnetic) phase. They are shown in *Figure 4*. Our simulation result agrees with the well-known first order phase transition in physics [7,8]. For the low noise level simulation, the final cluster state strongly depends on the initial configuration of shops. Thus our question of finding a good initial configuration is very meaningful and economically important. On the other hand, at high noise level, the final state evolved from any initial state is a randomized pattern, so that the discussion of what is the optimal initial configuration will be meaningless. This is the reason why we only analyse the low noise (T) case for the dependence of market share on initial configuration.

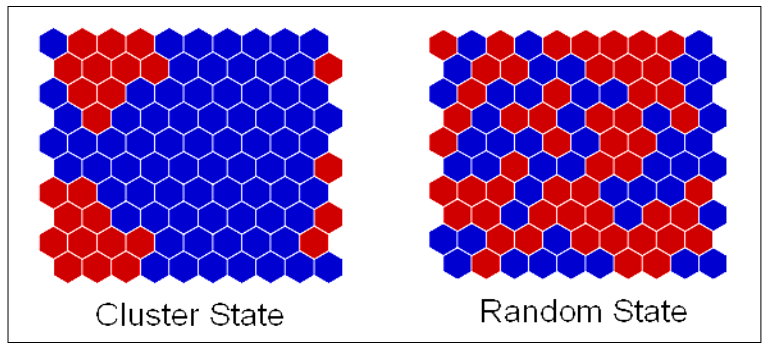

*Figure 4 Cluster State at low noise level and Random State at high noise level*

The optimal strategies for "blue" company at noise level = 1, 3, 4 are extracted. They are shown in *Figure 5*. We observe that all optimal configurations for "blue" company can gain more than 70% average market share after Monte Carlo simulation. Qualitatively, all the "blue" shops tend to form a big cluster so as to dominate the market. We further investigate the configuration by computing the connectivity distribution function. It is shown in *Figure 6*. The optimal configuration (Fig.5) usually has a Gaussian like distribution for the connectivity. We observe that a common feature among the winners, which are the blue distributions that evolve to dominate the market, in that their connectivity distributions are wider than the red ones.

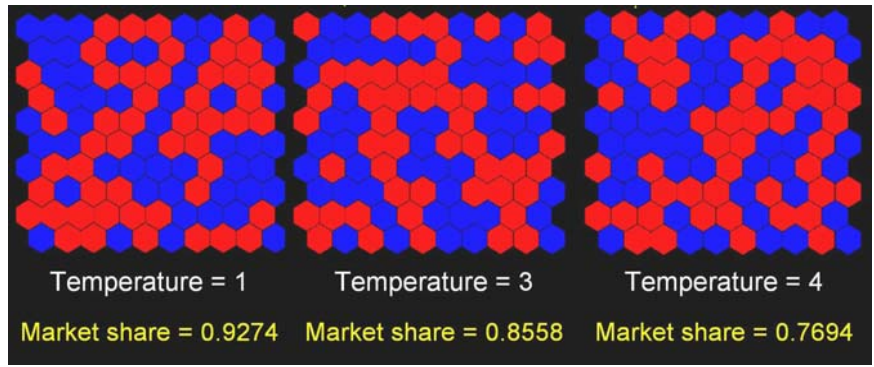

*Figure 5 The Optimal Shops Allocation and Final Average Market Share at Noise Level = 1, 3, 4*

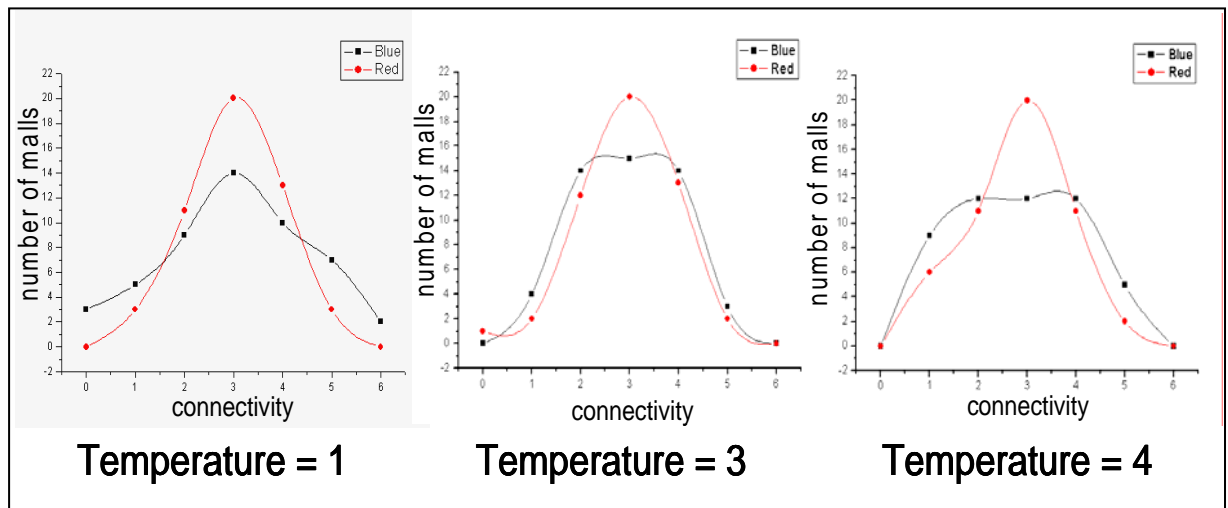

*Figure 6 The Connectivity Distribution at Noise Level = 1, 3, 4*

This observation is confirmed by fitting the connectivity distribution by a Gaussian and extract for the width. We find that the width of the connectivity distribution for the winner is bigger than the loser in all the noise level (T=1,3,4) tested. In order to gain a large market share, the administrator of a company should distribute his shops over a wide range of connectivity. Focusing on high connectivity alone will not work. Indeed, if all shops are connected with high connectivity, the effect of interaction between the blue shops and the red shops will be weak, resulting in a small chance for defeating the opponent. A general rule is to use part of the resource to form links between clusters, and the remaining resource be distributed to the cluster area of your opponent, so that your opponent cannot form a strong cluster. This also avoids wasting resource on some competitive sites that have 50% chances to be switched.

## Conclusion

The combined technique of Evolutionary algorithm and Monte Carlo simulation provides a good example for investigating the evolution of competitive market situation. Here we show the selection of good strategy for resource allocation (choosing good initial configuration) in the context of the Ising model on a hexagonal network, but our techniques can be easily extended to different lattice geometry and other models in statistical physics. Also, our observation of the connection between the distribution of connectivity and winning initial configurations demonstrates the power of topological analysis in the dynamics of evolving complex systems.

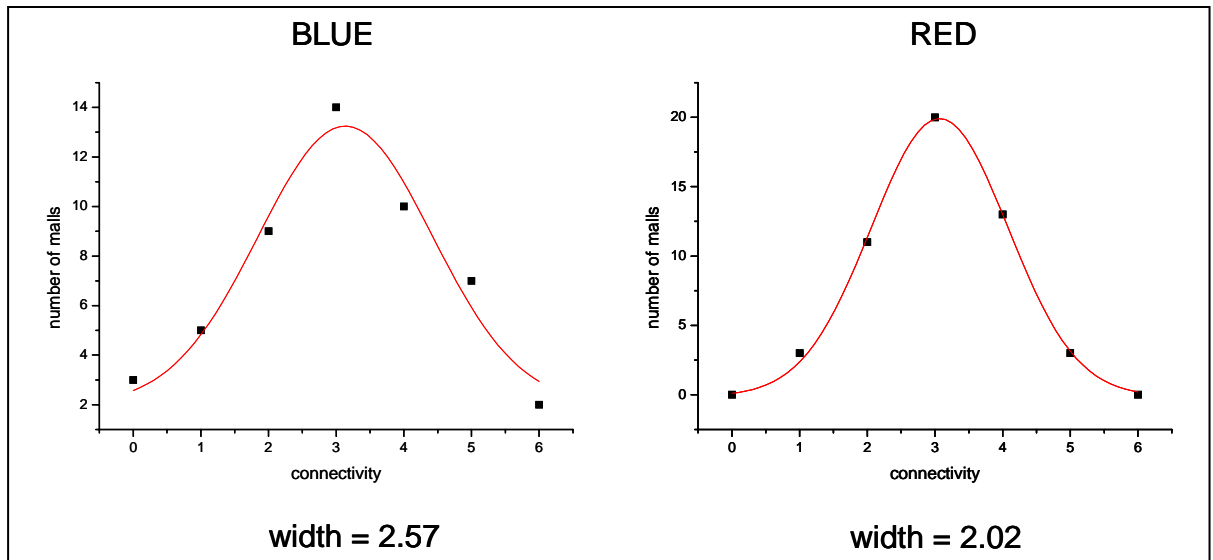

*Figure 7 The Connectivity Distribution fitted by a Gaussian function at Noise Level = 1*

## Acknowledgement

K.Y. Szeto acknowledges the support of grant CERG HKUST6157/01P. W.K. Cheung acknowledges fruitful discussion with Chun Wai Ma.